\documentclass[runningheads]{llncs}
\usepackage{graphicx}
\usepackage{booktabs}
\usepackage{multirow}
\usepackage{overpic}
\usepackage{color}
\usepackage{blindtext}
\usepackage{siunitx}
\usepackage{textcomp}
\usepackage{amsmath}

\begin{document}
%

%
\titlerunning{Breaking with Fixed Set Pathology Recognition}
%
%
%
%
%

\title{Breaking with Fixed Set Pathology Recognition through Report-Guided Contrastive Training}
\author{Constantin Seibold,\textsuperscript{\rm1} 
        Simon Rei\ss,\textsuperscript{\rm1}
        M. Saquib Sarfraz,\textsuperscript{\rm1}\\
        Rainer Stiefelhagen,\textsuperscript{\rm1}
        Jens Kleesiek\textsuperscript{\rm2}}
\authorrunning{Seibold et al.}
\institute{\textsuperscript{\rm 1} Karlsruhe Institute of Technology, Germany\\
\textsuperscript{\rm 2} University Medicine Essen, Germany\\
\textsuperscript{\rm 1}\{firstname.lastname\}@kit.edu,
\textsuperscript{\rm 2}\{firstname.lastname\}@uk-essen.de}
\maketitle              
\begin{abstract}
When reading images, radiologists generate text reports describing the findings therein. Current state-of-the-art computer-aided diagnosis tools utilize a fixed set of predefined categories automatically extracted from these medical reports for training. This form of supervision limits the potential usage of models as they are unable to pick up on anomalies outside of their predefined set, thus, making it a necessity to retrain the classifier with additional data when faced with novel classes. In contrast, we investigate direct text supervision to break away from this closed set assumption. By doing so, we avoid noisy label extraction via text classifiers and incorporate more contextual information.
    We employ a contrastive global-local dual-encoder architecture to learn concepts directly from unstructured medical reports while maintaining its ability to perform free form classification. 
    We investigate relevant properties of open set recognition for radiological data and propose a method to employ currently weakly annotated data into training.
    We evaluate our approach on the large-scale chest X-Ray datasets MIMIC-CXR, CheXpert, and ChestX-Ray14 for disease classification. We show that despite using unstructured medical report supervision, we perform on par with direct label supervision through a sophisticated inference setting. 
\end{abstract} 
\section{Introduction}
\setlength{\abovecaptionskip}{2pt}
\setlength{\belowcaptionskip}{-12pt}

\begin{figure*}[t]
    \centering
         \includegraphics[width=0.85\linewidth]{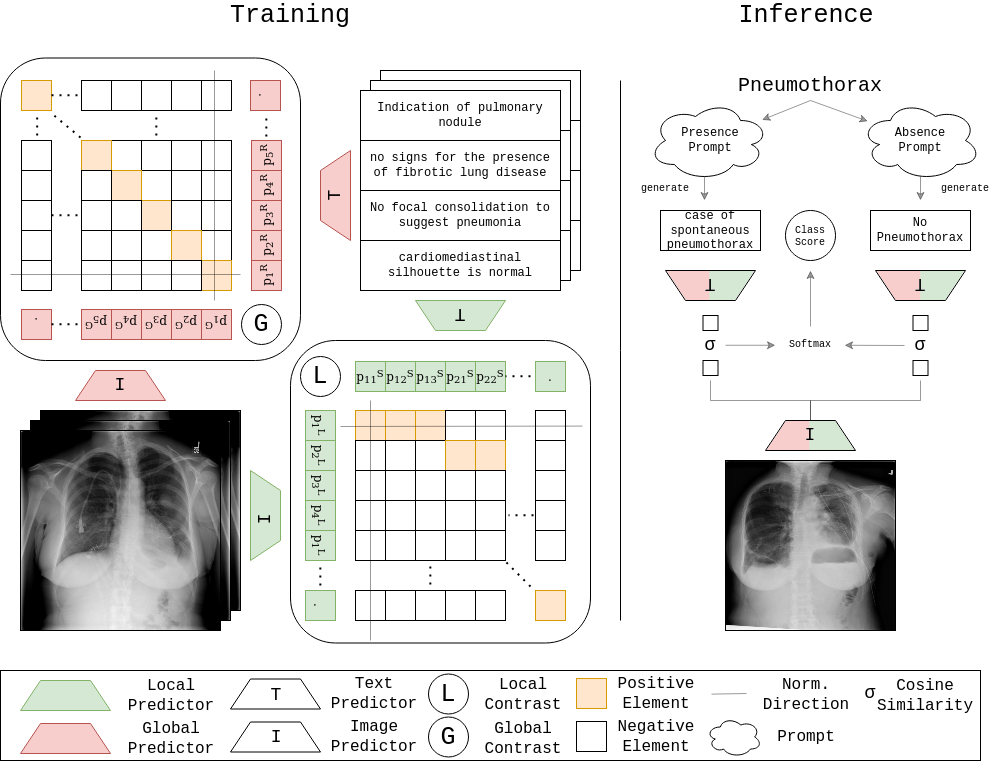}
    \caption{Illustration of our proposed method. Training on the left, inference on the right.}
    \label{fig:method}
\end{figure*}

Radiologists interpret a vast amount of imaging data and summarize their insights as medical reports. This documentation accumulates large databases of radiological imaging and accompanying findings, i.e.,  millions of collected chest radiographs annually~\cite{nhs}. Computer-aided-diagnosis (CAD) systems utilize these databases to streamline the clinical workflow and save time~\cite{qin2018computer,kim2019deep,jaiswal2019identifying}.
Modern CAD tools often rely on deep learning models~\cite{qin2018computer} using large-scale data sets such as MIMIC-CXR~\cite{wang2017chestx,johnson2019mimic,irvin2019chexpert,bustos2020padchest} for training.
Training for such tasks requires hand-designed supervision, typically by extracting a fixed set of predefined labels from the reports using rule- or deep learning-based models~\cite{wang2017chestx,irvin2019chexpert,smit2020chexbert}.
Such training typically requires hand-designed supervision in the form of extracting a set of labels from the reports using rule- or deep learning-based models~\cite{wang2017chestx,irvin2019chexpert,smit2020chexbert}.
While these tools can deliver acceptable performance on a subset of diseases~\cite{wu2020comparison}, they lack generalization capabilities for diseases that were not part of the fixed label-set used for training.
To add disease classes requires substantial effort annotating the data with extra labels and retraining the system. To circumvent this, one can approach training in a class-agnostic manner, however, it becomes unclear how models can still be applied to classify diseases.

Recent methods based on contrastive language-image pre-training~\cite{radford2021learning,jia2021scaling,pham2021combined,zhang2020contrastive,wang2021self} indicate that by large-scale multi-modal representation learning, object recognition can be detached from prior fixed-set, hand-designed class definitions. These models learn joint feature spaces between images and textual descriptions and utilize text prompts to transform recognition from learned fix-set classification to a matching task between text and image embeddings.
Radiological reports, in contrast to natural-image captions, have an inherently different structure, as they encompass multiple distinct sentences such that their entirety describes all relevant information. This shift makes a direct application of existing methods non-trivial. 

In this work, we see our contributions as the following: 
\begin{enumerate}
    \item We address training through report supervision by considering radiological reports in one of two ways: The local level, assuming each sentence conveys a distinct concept relevant for the patient, and secondly, the global report view, which encodes the entirety of the findings.
    \item We propose a novel inference setting that allows us to query any desired finding, and the CAD system generates a binary decision regarding its presence in the given radiological imagery.
    \item We provide an extensive study on various factors impacting the performance of multi-modal training and inference.
\end{enumerate}

\section{Global-Local Contrastive Learning}
We illustrate our method for report-based training of a vision model and inference protocol specifically designed for disease recognition in Figure \ref{fig:method}.
To tackle the complexity of medical reports, we split representations into a sentence- and report-level from a shared visual and language encoder. We consider embeddings for both the presence and absence of a pathology for its prediction. 
\subsection{Model Overview}
Contrastive language and image pretraining (CLIP) has shown immense potential for object recognition in natural images~\cite{radford2021learning} through learning from image-caption pairs. In contrast to textual descriptions in CLIP-based models, medical reports typically consists of multiple sentences focusing on different parts in the image. As each sentence contains specific subset of information, we aim to capture sentence- (local) and report-level (global)  context in our representations.
Thus, our model builds on separate image- and text encoders $\phi$ and $\theta$, which embed an image $I$ via $z_I = \phi(I)$ and a sentence by $z_s = \theta(s)$, respectively. 

In training, for a given report $R$, we capture the local context by splitting the full report into its sentences $R = \{s_1,\dots,s_n\}$ and subsequently extract sentence-level embeddings $z_{s_i} = \theta(s_i), s_i \in R$. To generate global embeddings that contain the full information of the whole report, the sentence-level embeddings are aggregated through attention pooling: $z_{R} = \text{Attn}([z_{s_1},\dots,z_{s_{|R|}}])$~\cite{NIPS2017_3f5ee243}.
To embed $z_I$, $z_s$ and $z_R$ into shared multi-modal representations, we project sentences and reports via linear transformations $p^S$ and $p^R$ into two feature spaces.
As the image encoder has access to global image information for report-level prediction as well as to local image patterns for selective sentence-dependent prediction, we project $z_I$ twice: into a global representation $p^{G}(z_I)$ which shall align with $p^R(z_R)$ and a representation for local patterns $p^{L}(z_I)$ for alignment with $p^S(z_s)$. 

During training, we are provided with a dataset of image-report pairs $(I_i, R_i) \in {(I_1, R_1), \dots, (I_N, R_N)}$.
For brevity and clarity in subsequent formulas, we will write projections, \textit{e.g.} the global projection of an image $I_i$ as $p^G_i$ instead of $p^G_i(\theta(I_i))$, for the projection of the $k^{th}$ sentence from report $R_i$, we write $p^S_{ik}$.

\subsection{Training Objectives}

\noindent\textbf{Local Contrast:} While radiological reports describe the assessment of a patient's health, not every sentence is directly linked to specific findings, some sentences mention clinical procedures or required follow-up examinations.
However, we can assume that all clinically relevant information is present in a subset of sentences in the report due to the doctors' obligation to document the findings.
This property is the core of the multiple-instance learning (MIL) assumption. 
Therefore, it might seem natural to choose MILNCE~\cite{miech2020end} as the MIL-based objective for integrate sentences in training, yet, this assumption only holds when normalizing over sentences, as not every sentence has to match the image. However, if a sentence fits an image, it should match strictly that image, thus, we hold the regular formulation when normalizing over images.
As such, MILNCE does not quite fit this use-case and we redesign it by splitting its symmetry:
\begin{equation}
    \begin{aligned}
        \mathcal{L}_{L}(I_i, R_i) =  & - \log \frac{
    \sum^{n}_{k=1} \exp(\sigma(p^L_i, p_{ik}^{S})/\tau_L) }
    {\sum^N_{j=1} \sum^{n}_{m=1}  \exp(\sigma(p_i^{L}, p_{jm}^{S}) /\tau_L)  } \\&- \sum^{n}_{k=1} log \frac{
    \exp(\sigma(p_i^{L}, p_{ik}^{S})/\tau_L) }
    {\sum^N_{j=1} \exp(\sigma(p_j^{L}, p_{ik}^{S}) /\tau_L)  },
    \end{aligned}
\end{equation}
with $\tau_L$ being a learned parameter and $\sigma(\cdot,\cdot)$ denoting the cosine similarity.

\noindent\textbf{Global Contrast:} For our batch we assume that an image-report pair is unique and formulate the following objective leveraging the attention-fused  reports via: 
\begin{equation}
    \mathcal{L}_{G}(I_i, R_i) =  - \log \frac{
    \exp(\sigma(p_i^{G}, p_{i}^{R})/\tau_G) }
    {\sum^N_{j=1} \exp(\sigma(p_i^{G}, p_{j}^{R}) /\tau_G)  } \\- log \frac{
    \exp(\sigma(p_i^{G}, p_{i}^{R})/\tau_G) }
    {\sum^N_{j=1} \exp(\sigma(p_j^{G}, p_{i}^{R}) /\tau_G)  }
\end{equation}

\noindent\textbf{Self-Supervision:} CLIP has been established as a data-hungry algorithm~\cite{radford2021learning,pham2021combined}. Several recent methods combine intrinsic supervision signals with the CLIP objective to make full use of the available data~\cite{li2021supervision,mu2021slip}. 
As we have access to severely smaller datasets in the medical domain as compared to the natural image domain, we follow Li~et~al.~\cite{li2021supervision} and integrate SimSiam~\cite{chen2021exploring}. For this, we generate two augmented versions of the input image $A_1(I)$ and $A_2(I)$ and add a three-layer encoder-head $p^{E}$ and a two-layer prediction-head $p^{P}$ on top of the visual backbone $\phi$ to enforce similarity between the two views:
\begin{equation}
    \mathcal{L}_{S}(A_1(I),A_2(I)) = - \sigma(p^{P}_{A_1(I)}, \text{detach}(p^{E}_{A_2(I)}))  - \sigma(p^{P}_{A_2(I)}, \text{detach}(p^{E}_{A_1(I)}))
\end{equation}
Furthermore, we utilize the augmented images used for the self-supervised objective to mirror our text-image objectives to the augmented samples. 
\begin{equation}
    \mathcal{L}_{M}(I_i,R_i) = \mathcal{L}_G(A_1(I_i), R_i) + \mathcal{L}_L(A_1(I_i), R_i) + \mathcal{L}_G(A_2(I_i), R_i) + \mathcal{L}_L(A_2(I_i), R_i)
\end{equation}

Our final objective for report-based contrastive learning amounts to: 
\begin{equation}
    \mathcal{L}(I_i,R_i) = 
    \lambda_1 * \mathcal{L}_{L}(I_i,R_i) +
    \lambda_2 * \mathcal{L}_{G}(I_i,R_i) +
    \lambda_3 * \mathcal{L}_{S}(I_i,R_i) +
    \lambda_4 * \mathcal{L}_{M}(I_i,R_i)
\end{equation}
with $\lambda_1 = \lambda_2 = \lambda_3 = 0.5$ and $\lambda_4 = 0.25$.

\subsection{Model Inference}
For fixed set classification models, the inference process is straightforward: A given image $I$ passed through a network with an activation in the final layer, returning class-wise pseudo probabilities. When model architecture and training procedure do not permit a classification layer, methods often resort to zero-shot-like inference~\cite{huang2021gloria,radford2021learning,wang2021self} where a nearest neighbor search in semantic space is conducted~\cite{frome2013devise}. 
In our considered design a text-based query is used to infer the presence or absence of a given disease. In similar CLIP-like models, the text embeddings (e.g., of the disease names) can be matched to an image embedding. The query with the maximum similarity can then be retained as matched.

Such a matching based disease discovery is feasible for detecting single disease class. The underlying assumption of having exclusively one dominant class to predict does not hold for chest radiographs as pathologies are not mutually exclusive. Similarly, modeling co-occurrence as individual classes is also infeasible due to the exponentially rising number of possible class combinations. As such, inference for a multi-label classification needs to be formulated for such contrastively trained methods.

We perform this, by querying an image with class-related textual prompts and interpreting their similarity scores as prediction probabilities for the respective disease class. In practice, we notice that a single query for class presence is ambiguous since the text embeddings of words and their negations may fall close to one another in the feature space. Due to this proximity of opposing semantics a query could be mistaken with the negation of its class. 

To overcome this issue, we propose to perform inference over two sets of queries ($q_c^p,q_c^n$) for each class. While the query embedding $q_c^p$  indicates the occurrence of a class $c$, $q_c^n$ indicates its absence, \textit{e.g.} \textsc{opacities consistent with pneumonia} for presence as opposed to \textsc{the lungs are clear} for its absence.
Then, the cosine similarity between the image and both queries $\sigma(p^I,q_c^p)$ and $\sigma(p^I,q_c^n)$ is computed with the final prediction wrt. class $c$ being defined as:
\begin{equation}
    P(c,I) = \frac{\exp(\sigma(p^I,q_c^p)/\tau)}{\exp(\sigma(p^I,q_c^p)/\tau) + \exp(\sigma(p^I,q_c^n)/\tau)},
\end{equation}
where $\tau$ is the respective learned scaling factor depending on the used projection.

\section{Prompt Engineering}
Several works on zero-shot classification perform their inference by extracting the features of the class name through a word embedding model~\cite{wang2019survey}. While sufficient for most zero-shot applications a lot of context regarding the class is lost.  In order to effectively utilize language-vision models it is necessary to align the downstream task to the training~\cite{radford2021learning}. 
As such we model a set of positive and negative prompts applicable for pathologies to enrich our matching process between visual and textual projections.
While for our basic approach we consider (`\{class\}', `No \{class\}') prompts, we found that a more detailed prompt design can overall deliver improved performance.  As such we consider a set of prompts following the templates (`\{adverb\}\{indication\_verb\} \{effect\}$^*$ \{location\}$^*$ \{class\_synonym\}', `\{adverb\} \{indication\_verb\} \{absence\} \{class\_synonym\}'). Hereby, we utilize all combinations over a small set of categories to gather a variety of different settings. During inference, features of all queries of the same set are averaged.  

\noindent\textbf{Prompt-based Dataset Extension: } Despite medical reports being the more common resource in the practical field, currently the majority of large-scale datasets are only publically available with fixed sets of labels. In order to investigate the effect of additional data in training of our method, we reverse our proposed prompt engineering to generate synthetic reports for the datasets PadChest and ChestX-Ray14 based on their class-labels.
Through this procedure, we are able to sample sentences indicating presence or absence of a class and generate more than 200k added image-report pairs.

\begin{table}[b]
    \centering
    \begin{tabular}{l@{\hspace{0.2cm}}r}
        \begin{tabular}{lcccccccc}
        \toprule
          Inf. & MIMIC   & CheXpert  & CXR14 & Avg.\\  
          \midrule
           Local & 77.81 & 78.09 & 71.72 & 75.87\\
           Global & 76.24 & 80.42 &   71.00 & 75.88\\
           \midrule
           Max  & 76.85 & 71.29 & 78.22 & 75.45\\
           Cat  & 77.29 & 80.30 & 71.72 & 76.43 \\
           Mean & 77.06  & 81.08 &  71.50 &  76.54 \\
          \bottomrule
        \end{tabular}
         &  
         \begin{tabular}{lcccccccc}
            \toprule
            
            Parts & MIMIC   & CXpert  & CXR14 & Avg.\\ 
              \midrule
               $\mathcal{L}_G$ & 75.47 & 77.24 & 69.22 & 73.97 \\ 
               $\mathcal{L}_G$+$\mathcal{L}_L$ & 76.20 & 82.24 & 69.26  &   75.90 \\ 
               $\mathcal{L}_G$+$\mathcal{L}_L$+$\mathcal{L}_S$ & 76.10 & 76.08 & 74.24 & 75.47 \\ 
               $\mathcal{L}_G$+$\mathcal{L}_L$+$\mathcal{L}_M$  & 77.03 & 77.36 & 71.72 & 75.37\\  
               \midrule
               Ours & 77.06 & 81.08 & 71.50 & 76.54 \\ 
              \bottomrule
            \end{tabular}
    \end{tabular}

    \caption{Left: Impact of chosen scores for inference. Right: Ablation of model parts.}
    \label{tab:head}
\end{table}

\begin{figure*}[t]
    \centering
    \begin{tabular}{cc}
       \includegraphics[width=0.65\linewidth,height=0.275\linewidth]{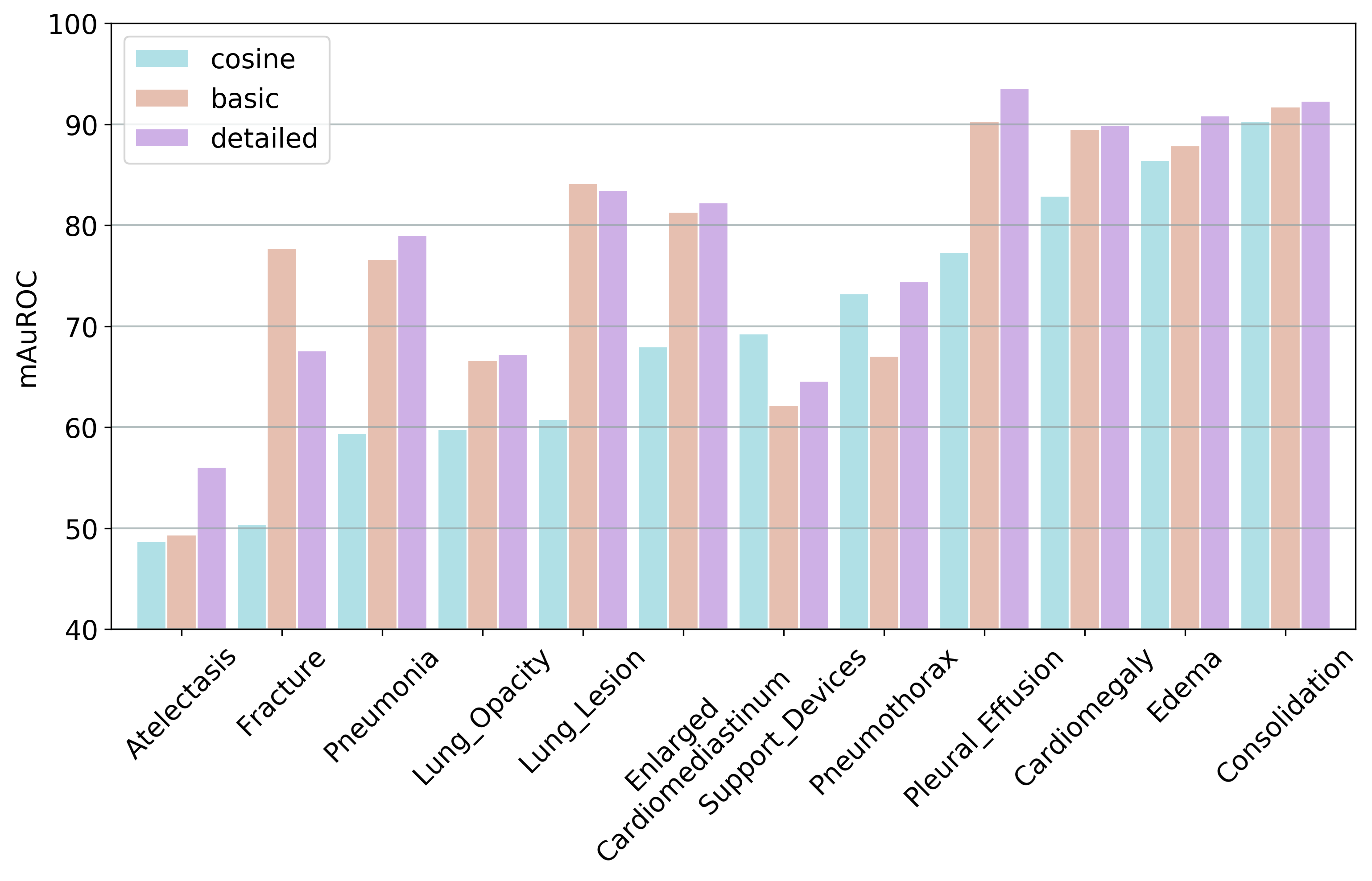}  &  \includegraphics[width=0.3\linewidth,height=0.275\linewidth]{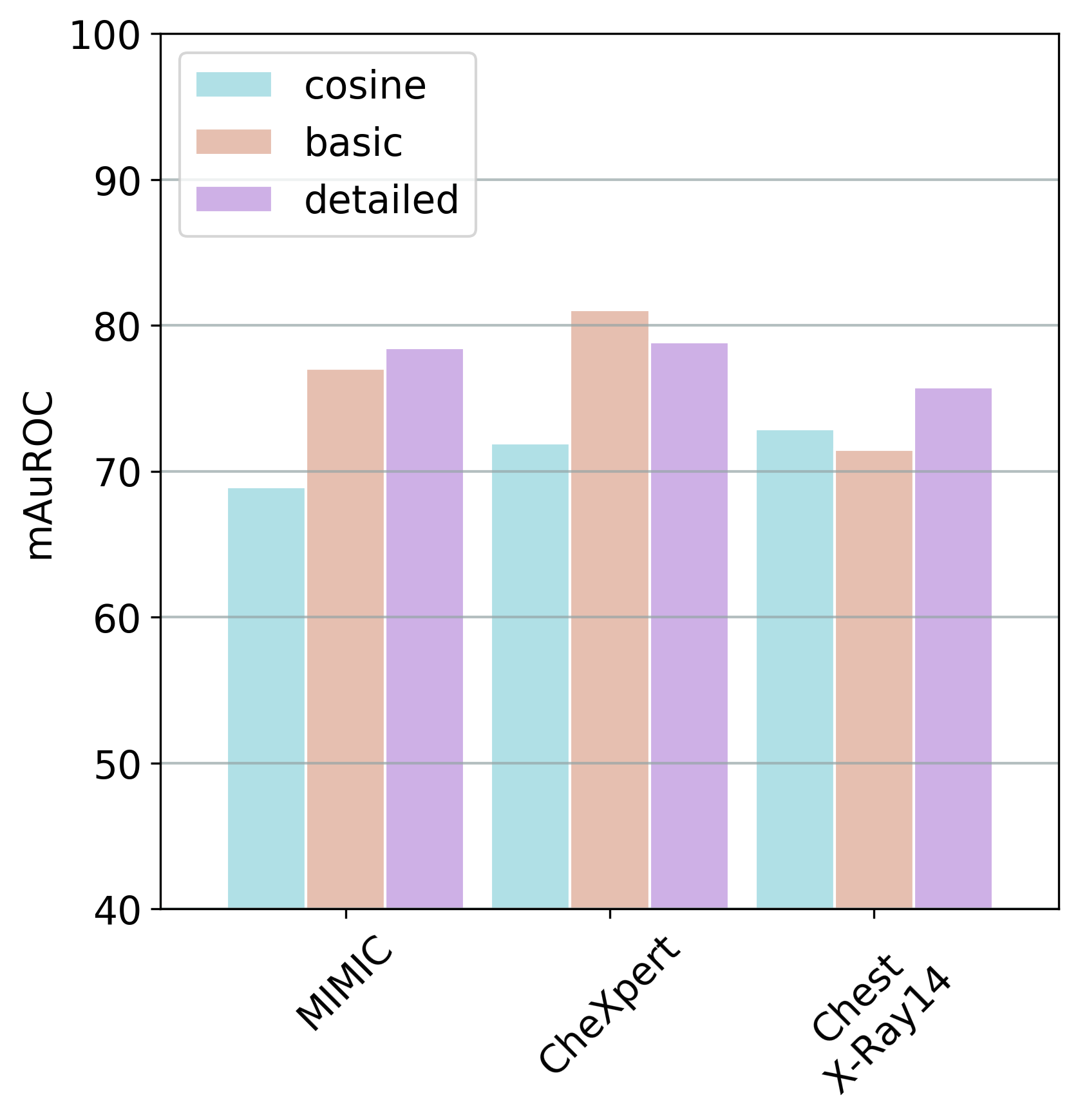}
    \end{tabular}
    
    \caption{Performance changes based on differences in prompt generation. Class wise performance on the left. Mean performance to the right. Models trained on MIMIC.}
    \label{fig:prompts}
\end{figure*}

\section{Experiments}
\subsection{Experimental Setup}
\noindent\textbf{Datasets:}
\begin{itemize}
    \item \textbf{MIMIC-CXR: } It contains 377,110 
        chest X-rays taken from 65179 patients with 14 disease labels and 227,835 reports.
        We use the splits provided by~\cite{johnson2019mimic}. Unless further specified all models were trained on this dataset. 
\item \textbf{CheXpert: } It contains 224,316
    chest X-rays taken from 65,240 patients with 14 disease labels. The labels are shared with MIMIC-CXR.
    We only consider the validation split provided by~\cite{irvin2019chexpert}.
\item \textbf{ChestX-ray14: } It contains 112,120
frontal-view chest X-rays taken from 30,805 patients with 14 disease labels.
    We use the splits provided by~\cite{seibold2020self}.
    \item \textbf{PadChest: } It consists  160k
    chest X-rays of 67k patients with 174 findings.
\end{itemize}
\noindent\textbf{Evaluation Setup: } We evaluate the multi-label classification ability of all networks via the Area Under the ROC-curve (AUROC) and show the performance over MIMIC-CXR, CheXpert and ChestX-Ray14. For all experiments expect Table~\ref{tab:final} we consider validation performance. Labels with value -2 and -1 are ignored for the calculation of the metric as their state is not certain. For all ablations, we use the "basic"-prompting scheme, while for further experiments the "detailed"-scheme is used. 

\noindent\textbf{Implementation Details: } For all experiments we use the same ResNet50 and Transformer as Redford~et~al.~\cite{radford2021learning} as backbones. We optimize with AdamW~\cite{loshchilov2017decoupled}, a learning rate of 0.0001 and a cosine schedule. We trained classification models with a learning rate of 0.0005 as this has shown slightly better performance. During training, we resize the images to the inference size of 320 × 320 and randomly crop by 288x288. For specific further augmentations, we follow SimSiam~\cite{chen2021exploring}.

\subsection{Results}
\noindent\textbf{Ablation - Effect of Heads:}
We investigate the impact of both prediction heads during inference. We start by showing the individual head performance and then go over to different fusion approaches on the left of Table\ref{tab:head}. For feature fusion we consider the concatenation of local and global features of the same modality. For score-fusions, we calculate scores as described above and aggregate the class predictions based on their maximum or average.

We see that for our method both the global and local head show nearly the same performance. While performing max-score fusion the across-dataset-performance drops by 0.4\%, where mean-score fusion improves by 0.6\%.
\begin{figure*}[t]
    \centering
    \begin{tabular}{cc}
       \includegraphics[width=0.65\linewidth,height=0.3\linewidth]{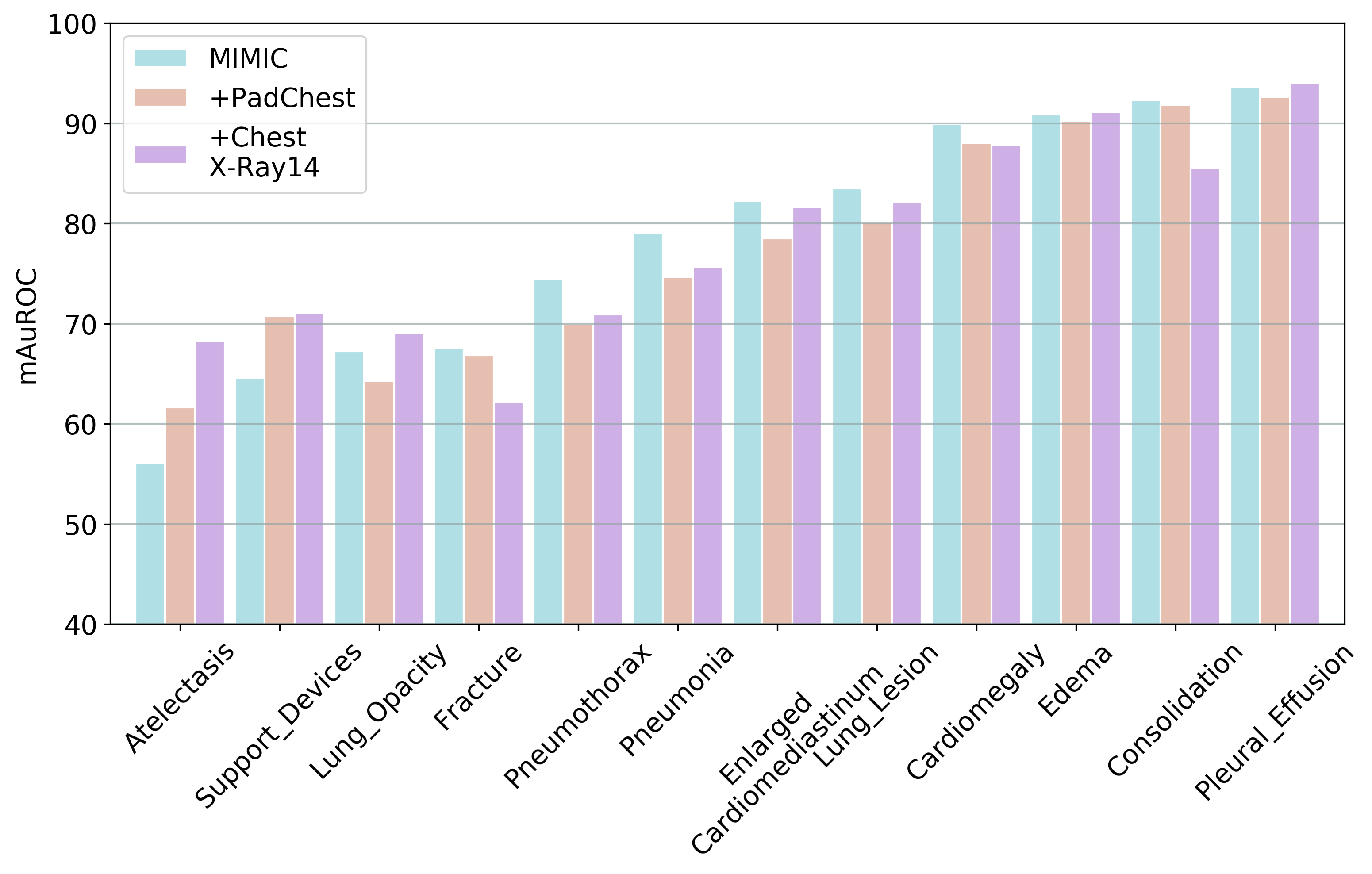}  &  \includegraphics[width=0.3\linewidth,height=0.3\linewidth]{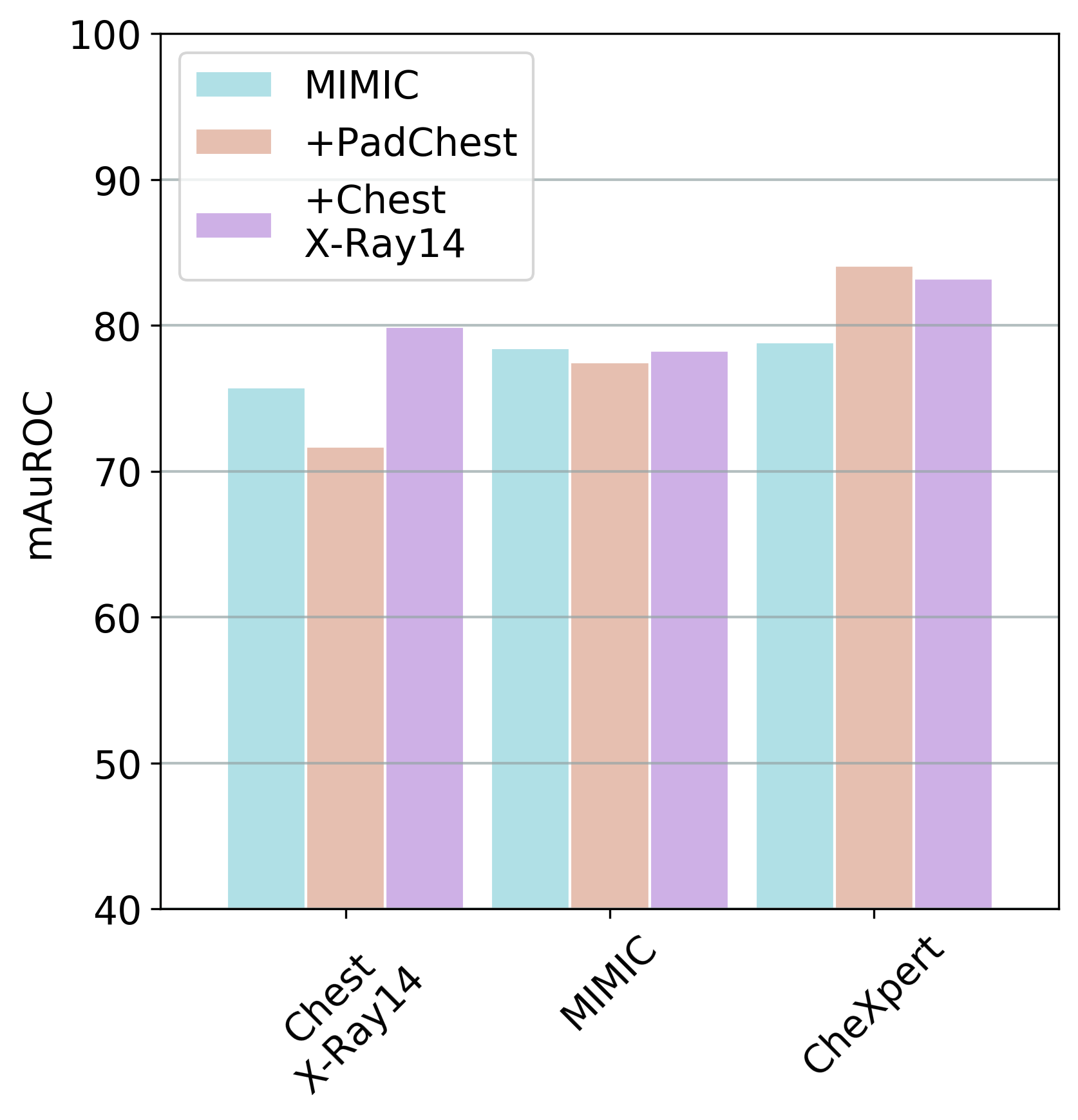}
    \end{tabular}
    
    \caption{Contributions of data scaling for chest radiograph dataset.  Performance change of adding additional chest X-ray datasets with prompt-based captions.}
    \label{fig:data}
\end{figure*}
\begin{table}[h]
    \centering
    \begin{tabular}{lccccccccc}
    \toprule
    \multirow{3}{*}{Method} & & \multicolumn{2}{c}{MIMIC-CXR }  && \multicolumn{1}{c}{CheXpert } && \multicolumn{2}{c}{ChestX-Ray14 } \\
    & &  \multicolumn{2}{c}{(in-domain)} &&  \multicolumn{1}{c}{(out-of-domain)} &&  \multicolumn{2}{c}{(out-of-domain)} \\ 
    & &  val & test &&   val &&  val & test \\ 
    
    \midrule
    Label-Supervised & & 77.26 & 77.42 & & 78.90 & & 79.70 &  76.47  \\
    \midrule
    CLIP                          & & 73.23 & 70.25 & & 75.85 & & 68.03 & 63.34  \\
    SLIP                          & & 72.45 & 72.44 & & 78.49 & & 71.45 & 67.55  \\
    \midrule
    $MILNCE_{local}$                & & 69.30 & 69.18 & & 74.98 & & 67.56 & 63.06 \\
    LoCo                          & & 77.03 & 78.15 & & 81.71 & & 71.92 & 68.14  \\
    GloCo                         & & 75.47 & 76.58 & & 77.24 & & 69.22 & 65.86  \\
    Ours                          & & 78.46 & 79.40 & & 78.86 & & 75.77 & 71.23  \\
    \midrule
    Ours*                         & & 78.30 & 80.40 & & 83.24 & & 79.90 & 78.33  \\
     \bottomrule
    \end{tabular}
    \caption{Classification performance on MIMIC, CheXpert and Chest-XRay14. * indicates that the model was trained with additional PadChest and ChestX-Ray14 data.}
    \label{tab:final}
\end{table}

\noindent\textbf{Ablation - Effect of Losses:}
On the right of Table\ref{tab:head}, we show the impact of the objective functions. We see that adding local contrast improves the model by 2\%. Adding the self-supervised and mirrored objective worsen performance by 0.45\%. It can be noted that the self-supervised loss achieves the best performance on ChestX-Ray14 by more than 2\%.
 whereas adding both simultaneously improves across dataset performance by 0.6\%. 

\noindent\textbf{Multi-label Inference and Prompt Engineering:} We show the impact of our proposed inference scheme and prompts in Figure~\ref{fig:prompts}. We see that performance overall improves with significant improvements for some classes such as fractures which were unable to be categorized just using cosine-similarity. When the detailed prompt the mean performance further improves.

\noindent\textbf{Data size Impact: }
 We show the impact of using additional prompt-based reports during training  in Figure~\ref{fig:data}. We see that including artificial training data for ChestX-Ray14 significantly improves its validation performance.  In general it seems that while for some classes performance seems to worsen, the overall performance improves when adding additional data.

\noindent\textbf{Comparison with Other Approaches: }
We compare against the same vision network trained with label supervision on its respective dataset. All other methods were trained using the MIMIC-CXR dataset. SLIP~\cite{mu2021slip} refers to a version of CLIP, which incorporates self-supervision in form of a SIMCLR-like objective~\cite{chen2020simple}. $MILNCE_{local}$ refers to our local branch trained with the MILNCE~\cite{miech2020end} objective alone. LoCo and GloCo refer to our method trained with either just the local or global objective respectively. We evaluate using the "detailed"-prompt scheme.  We show the results in Table~\ref{tab:final}. 

We see that our formulation of the local contrastive loss outperforms the MILNCE version across all datasets. Our proposed method outperforms the considered contrastive language-image pretraining baselines in the form of CLIP and SLIP and manages to achieve similar performance as the supervised ResNet for domains similar to MIMIC, however, underperforms for the ChestX-Ray14 dataset. When adding the additional report datasets of PadChest and ChestX-Ray14 we manage to beat label-supervised performance across all datasets.

\section{Conclusion}
In this paper, we proposed an approach to make networks less reliant to label supervision through contrastive language-image pre-training on report level. In order to still maintain competitive levels of performance we introduced a novel way of constructing inference. Doing so we are able to offset issues stemming from explicit class similarities. 
We show that despite using unstructured medical report supervision, we perform on par with explicit label supervision through a sophisticated inference setting across different datasets. 

\section{Acknowledgements}
The present contribution is supported by the Helmholtz Association under the joint research school “HIDSS4Health – Helmholtz Information and Data Science School for Health” and by the Helmholtz Association Initiative and
Networking Fund on the HAICORE@KIT partition.
\bibliographystyle{splncs04}
\bibliography{miccai22}

\end{document}